\def\year{2020}\relax
\documentclass[letterpaper]{article} 
\usepackage{aaai20}  
\usepackage{times}  
\usepackage{helvet} 
\usepackage{courier}  
\usepackage[hyphens]{url}  
\usepackage{amsmath}
\usepackage{mathtools}
\usepackage{amsthm}
\usepackage{nicefrac}
\usepackage[symbol]{footmisc}
\usepackage{amssymb}
\usepackage{enumitem}
\usepackage{graphicx} 
\def\UrlFont{\rm}  
\usepackage{graphicx}  
\usepackage{algorithm}
\usepackage{algorithmic}
\usepackage[linesnumbered, ruled, vlined, algo2e, ruled]{algorithm2e}
\newtheorem{theorem}{Theorem}
\newtheorem{prop}{Proposition}
\newtheorem{innercustomgeneric}{\customgenericname}
\providecommand{\customgenericname}{}
\newcommand{\newcustomtheorem}[2]{%
  \newenvironment{#1}[1]
  {%
   \renewcommand\customgenericname{#2}%
   \renewcommand\theinnercustomgeneric{##1}%
   \innercustomgeneric
  }
  {\endinnercustomgeneric}
}
\newcustomtheorem{customthm}{Theorem}
\newcommand{\Var}{\mathrm{Var}}
\newcommand{\Cov}{\mathrm{Cov}}

\nocopyright
\title{Particle Smoothing Variational Objectives}
\author{
Antonio Khalil Moretti,\thanks{Equal contribution}\textsuperscript{\rm 1} Zizhao Wang,\footnotemark[1]\textsuperscript{\rm 1} Luhuan Wu,\footnotemark[1]\textsuperscript{\rm 2} Iddo Drori,\textsuperscript{\rm 1, 3} Itsik Pe'er \textsuperscript{\rm 1} \\ 
\textsuperscript{\rm 1} Department of Computer Science, Columbia University\\
\textsuperscript{\rm 2} Data Science Institute, Columbia University\\
\textsuperscript{\rm 3} School of Operations Research and Information Engineering, Cornell University\\
amoretti@cs.columbia.edu, z2504@columbia.edu, lw2827@columbia.edu,\\ idrori@cs.columbia.edu, itsik@cs.columbia.edu
}
 
\begin{document}

\maketitle

\begin{abstract}
A body of recent work has focused on constructing a variational family of filtered distributions using Sequential Monte Carlo (SMC). Inspired by this work, we introduce Particle Smoothing Variational Objectives (SVO), a novel backward simulation technique and smoothed approximate posterior defined through a subsampling process. SVO augments support of the proposal and boosts particle diversity. 
Recent literature argues that increasing the number of samples $K$ to obtain tighter variational bounds may hurt the proposal learning, due to a signal-to-noise ratio (SNR) of gradient estimators decreasing at the rate $\mathcal{O}(\sqrt{1/K})$. 
As a second contribution, we develop theoretical and empirical analysis of the SNR in filtering SMC, which motivates our choice of biased gradient estimators. We prove that introducing bias by dropping \textsc{Categorical} terms from the gradient estimate or using Gumbel-Softmax mitigates the adverse effect on the SNR. We apply SVO to three nonlinear latent dynamics tasks and provide statistics to rigorously quantify the predictions of filtered and smoothed objectives. SVO consistently outperforms filtered objectives when given fewer Monte Carlo samples on three nonlinear systems of increasing complexity.
\end{abstract}

\section{Introduction}
\noindent 
Sequential data can often be understood as a set of ordered, discrete-time measurements taken on a hidden dynamical system.
In the past years, there has been a large body of work concerned with inferring both the latent trajectories and dynamics of these systems when either the evolution or the observations are nonlinear~\cite{archer,deepkalman,lfads2,vind2,moretti2019smoothing}. 
Variational Inference (VI) and Markov Chain Monte Carlo (MCMC) are two popular such approaches. Connections between them have been
established recently, specifically by defining a flexible variational family of 
distributions using Sequential Monte Carlo (SMC)  
~\cite{anh2018autoencoding,NIPS2017_7235,pmlr-v84-naesseth18a}. These distributions are \emph{filtered} in the sense that the estimate of the latent state at time $t$ makes use only of information collected in the observations up to that time. 
The three main contributions of this paper:
\begin{itemize}
    \item \textbf{Particle Smoothing Variational Objective}:
    We propose an SMC method to construct \emph{smoothed} variational objectives (SVOs), that is, with the estimate for the latent state and dynamics conditioned on the full time ordered sequence of observations. We introduce a novel recursive backward-sampling algorithm and approximate posterior defined through a subsampling process. This augments the support of the proposal and boosts particle diversity.
    We further show that SVO is a well-motivated evidence lower bound by proving the unbiasedness of target in the extended particle space.
To quantify the learned dynamics, we repeatedly apply the transition function in the target to propagate the system forwards without input data and then use the emission function to make observation predictions. We show that our smoothed objective generates an improved estimate of the latent states as measured by the ability of the target to more accurately predict observations using the dynamics learned. 

\item \textbf{SNR Guarantees}:
Recent literature argues that increasing the number of samples $K$ to obtain tighter variational bounds may hurt the proposal learning, due to a signal-to-noise ratio (SNR) of gradient estimators decreasing at the rate $\mathcal{O}(\sqrt{1/K})$ \cite{rainforth2018tighter}.  In \cite{anh2018autoencoding} it was speculated that a result similar to \cite{rainforth2018tighter} holds for filtering SMC, motivating the design of distinct variational bounds for generative and proposal networks. SMCs resampling step introduces challenges for standard reparameterization due to the \textsc{Categorical} distribution. As a second contribution, we analyze the SNR for filtering SMC. 
We prove that SNR degradation does not apply to the inference network of filtering SMC due to the resampling step. We present theoretical and empirical evidence pointing to 
an increasing SNR dependent on the choice of 
the gradient estimator.

\item \textbf{Applications}: We apply SVO to two benchmark latent nonlinear dynamical systems tasks and single cell electrophysiology data from the Allen Institute~\cite{allennature}. 
SVO consistently outperforms filtered objectives when given fewer Monte Carlo samples.
\end{itemize}

\section{Preliminaries}
\paragraph{Inference in State Space Models}

Let $\mathbf{X} \equiv \{\mathbf{x}_1,\dots\mathbf{x}_T\}$ denote a sequence of $T$ observations of a $\mathbb{R}^{d_\mathbf{x}}$-dependent random variable. State space models (SSMs) propose a generating process for $\mathbf{X}$ through a sequence $\mathbf{Z}\equiv\{\mathbf{z}_1,\dots\mathbf{z}_T\}$, $\mathbf{z}_t\in \mathbb{R}^{d_\mathbf{z}}$ of unobserved latent variables, that evolves according to a stochastic dynamical rule. The joint density then satisfies:
\begin{equation}
p_{\theta}(\mathbf{X},\mathbf{Z}) =  F_\theta(\mathbf{Z}) \cdot  \prod_{t=1}^T g_\theta(\mathbf{x}_t|\mathbf{z}_t) \,, \label{eq:mgen}
\end{equation}
where  $g_\theta(\mathbf{x}|\mathbf{z})$ is an observation model, and $F_\theta(\mathbf{Z})$ is a prior representing the evolution in the latent space. In this work, we focus on the case of Markov evolution with Gaussian conditionals:
\begin{equation}
  F_\theta(\mathbf{Z})  = f_1(\mathbf{z}_1) \prod_{t=2}^T f_\theta(\mathbf{z}_t | \mathbf{z}_{t-1})\,, \nonumber\label{F}
\end{equation}
\begin{equation}
  f_1  = \mathcal{N}\big( \psi_1, \mathbf{Q}_1 \big) \,, \qquad \mathbf{z}_t  \sim \mathcal{N}\big( \psi_\theta(\mathbf{z}_{t-1}),\, \mathbf{Q} \big) \,. \label{z|z}
\end{equation}
Inference in SSMs requires marginalizing the joint distribution with respect to the hidden variables $\mathbf{Z}$,
\begin{equation}
\log p_\theta(\mathbf{X}) =  \int \log p_\theta(\mathbf{X}, \mathbf{Z}) \, d\mathbf{Z}.
\end{equation}
This procedure is intractable when 
$\psi_\theta(\mathbf{z}_{t})$ is a nonlinear function or when $g_\theta(\mathbf{x}_t|\mathbf{z}_t)$ is non-Gaussian.

\paragraph{Variational Inference} VI describes a family of techniques for approximating $\log p_\theta(\mathbf{X})$ when marginalization is analytically impossible. The idea is to define a tractable distribution $q_\phi(\mathbf{Z}|\mathbf{X})$ and then optimize a lower bound to the log-likelihood:
\begin{equation}
\log p_\theta(\mathbf{X}) \geq \mathcal{L}_{\text{ELBO}}(\theta,\phi, \mathbf{X}, \mathbf{Z}) = \underset{q}{\mathbb{E}}\Bigg[\log \frac{p_\theta(\mathbf{X}, \mathbf{Z})} {q_\phi(\mathbf{Z}|\mathbf{X})}\Bigg] \,. \label{ELBO}
\end{equation}
Tractability and expressiveness of the variational approximation $q_\phi(\mathbf{Z}|\mathbf{X})$ are contrasting goals. 
Auto Encoding Variational Bayes~\cite{DBLP:journals/corr/KingmaW13} (AEVB) is a method to simultaneously 
train 
$q_\phi(\mathbf{Z}|\mathbf{X})$ and $p_\theta(\mathbf{X},\mathbf{Z})$. 
The expectation value in Eq. (\ref{ELBO}) is approximated by summing over samples from the recognition distribution; which in turn are drawn by evaluating a deterministic function of a $\phi$-independent random variable (the reparameterization trick). 
Building upon this, the Importance Weighted Auto Encoder~\cite{journals/corr/BurdaGS15,NIPS2018_7699} (IWAE) constructs tighter bounds than the AEVB through mode averaging as opposed to mode matching. The idea to achieve a better estimate of the log-likelihood is to draw $K$ samples from the proposal and to average probability ratios.

\paragraph{Filtering SMC}
SMC is a family of techniques for inference in SSMs with an intractable joint. Given a proposal distribution $q_{\phi}(\mathbf{Z}|\mathbf{X})$ 
these methods operate sequentially, approximating
$p_{\theta}(\mathbf{z}_{1:t}| \mathbf{x}_{1:t})$ (the \textit{target}) for each $t$ by performing inference on a sequence of increasing probability spaces. 
$K$ samples (\textit{particles}) are drawn from a proposal distribution and used to compute importance weights:
\begin{align}
\mathbf{z}_t^{k} \sim q_\phi(\mathbf{z}_t^{k}|\mathbf{z}_{t-1}^{k},\mathbf{x}_t)\ , \quad w_t^{k} \coloneqq \frac{f_\theta(\mathbf{z}_t^{k}|\mathbf{z}_{t-1}^{k})g_\theta(\mathbf{x}_t|\mathbf{z}_t^{k})}{q_\phi(\mathbf{z}_t^{k}|\mathbf{z}_{t-1}^{k},\mathbf{x}_t)}.
\end{align}
A resampling strategy ensures that particles remain on regions of high probability mass. SMC accomplishes this goal by resampling the particle indices (\textit{ancestors}) according to their weights at the previous time step:
\begin{align}
a_{t-1}^{k} &\sim \textsc{Categorical}(\cdot|\bar{w}_{t-1}^{1},\cdots,\bar{w}_{t-1}^K)\ ,  \\
w_t^{k} &\coloneqq \frac{f_\theta(\mathbf{z}_t^{k}|\mathbf{z}_{t-1}^{a_{t-1}^{k}})g_\theta(\mathbf{x}_t|\mathbf{z}_t^{a_{t-1}^{k}})}{q_\phi(\mathbf{z}_t^{{k}}|\mathbf{z}_{t-1}^{a_{t-1}^{k}},\mathbf{x}_t)}. \nonumber
\end{align}
The posterior can be evaluated at the final time step. The functional integral is approximated below where $\delta_{\mathbf{z}_{1:T}^{k}}(\mathbf{z}_{1:T})$ is the Dirac measure:
\begin{equation}
\sum\limits_{k=1}^{K}\bar w_T^{k}\delta_{\mathbf{z}_{1:T}^{k}}(\mathbf{z}_{1:T}) \quad \text{where}\quad  \bar w_T^{k} = w_T^{k}/\sum_{j=1}^{K} w_T^{j}.
\end{equation}
The SMC algorithm is deterministic conditioning on $(\mathbf{z}_{1:T}^{1:K}, a_{1:T-1}^{1:K})$ \cite{NIPS2017_7235,anh2018autoencoding}. This implies that the proposal density can be reparameterized to act as a variational distribution that can be encoded:
\begin{align}
 &Q_{\text{SMC}}
 (\mathbf{Z}_{1:T}^{1:K}, \mathbf{A}_{1:T-1}^{1:K}) \coloneqq \Bigg(\prod\limits_{k=1}^{K}q_{1,\phi}(\mathbf{z}_1^{k}) \Bigg) \\
 &\times \prod\limits_{t=2}^{T}\prod\limits_{k=1}^{K}q_{t,\phi}(\mathbf{z}_t^{k}|\mathbf{z}_{1:t-1}^{a_{t-1}^{k}})\cdot \textsc{Categorical}(a_{t-1}^{k} |\bar{w}_{t-1}^{1:K}). \nonumber
  \label{eq:ELBO_SMC}
 \end{align}
An unbiased estimate for the marginal likelihood and the corresponding objective are defined below:
\begin{equation}
\hat{\mathcal{Z}}_{\text{SMC}} \coloneqq \prod\limits_{t=1}^{T}\Big[\frac{1}{K}\sum\limits_{k=1}^{K}w_t^{k} \Big],  \quad \mathcal{L}_{\text{SMC}} \coloneqq \underset{Q_{\text{SMC}}}{\mathbb{E}} \left[\log \hat{Z}_{\text{SMC}} \right].
\end{equation}

 \begin{figure*}
\centering
\quad
\includegraphics[width=0.85\textwidth]{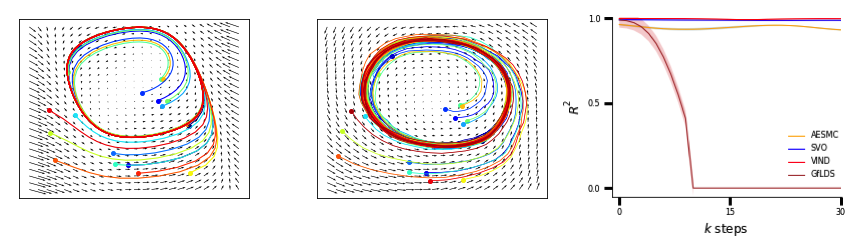}
\caption{Summary of the Fitzhugh-Nagumo results: the observation is one-dimensonal while the phase space and latent variables are two-dimensional; (left) ground truth dynamics and trajectories for the original system;  (center) latent dynamics and trajectories inferred by SVO; Initial points (denoted by markers) located both inside and outside the limit cycle are topologically invariant in the SVO reconstruction; (right) $R^2_k$ for various models on the dimensionality expansion task. Results are averaged over 3 random seeds.}
\label{fhn}
\end{figure*}

\paragraph{Particle Smoothing with Backward Simulation}
Forward Filtering Backward Simulation (FFBSi) ~\cite{doi:10.1198/016214504000000151} is an approach 
to approximate the smoothing posterior which admits the following factorization
\begin{equation}
p(\mathbf{z}_{1:T}|\mathbf{x}_{1:T}) = p(\mathbf{z}_T|\mathbf{x}_{1:T})\prod\limits_{t=1}^{T-1}p(\mathbf{z}_t|\mathbf{z}_{t+1:T},\mathbf{x}_{1:T})\ ,
\label{eq:smoothingposterior}
\end{equation}
where, by Markovian assumptions, the conditional backward kernel can be written as:
\begin{equation}
p(\mathbf{z}_t|\mathbf{z}_{t+1},\mathbf{x}_{1:T}) \propto p(\mathbf{z}_t|\mathbf{x}_{1:t})f(\mathbf{z}_{t+1}|\mathbf{z}_t).
\end{equation}
FFBSi begins with filtering to obtain $\{\mathbf{z}_{1:T}^{1:K}, w_{1:T}^{1:K}\}$ which provides a particulate approximation to the backward kernel:
\begin{align}
p(\mathbf{z}_t | \mathbf{z}_{t+1}, \mathbf{x}_{1:T}) &\approx \sum_{i=1}^K w_{t|t+1}^k \delta_{\mathbf{z}_t^i} (\mathbf{z}_t), \\ \text{where} \qquad \ w_{t|t+1}^i &= \frac{w_t^i f(\mathbf{z}_{t+1}|\mathbf{z}_t^i)}{\sum\limits_{j=1}^{K}w_t^j f(\mathbf{z}_{t+1}|\mathbf{z}_t^j)}. \nonumber
\end{align}
Backward simulation generates states in the reverse-time direction conditioning on future states by choosing $\tilde{\mathbf{z}}_t = \mathbf{z}_t^i$ with probability $w_{t|T}^i$. This corresponds to a \textit{discrete} resampling step in the backward pass.
As a result the backward kernel is approximated from particles that are drawn from the proposal $q(\mathbf{z}_t|\mathbf{z}_{t-1})$ in the forward pass.

\section{Particle Smoothing Variational Objectives}
We will utilize the smoothing posterior in Eq. (\ref{eq:smoothingposterior}) to define a backward proposal distribution and sample trajectories to construct a variational objective. We propose a novel approximate posterior to overcome the limitation of the FFBSi by augmenting the support of the backward kernel through the subsampling of auxiliary random variables.

\paragraph{Overview}
 We provide an overview of Particle Smoothing Variational Objectives (SVO) before presenting a detailed derivation and description in Algorithm~\ref{alg:svoFFBSi} (we have annotated the overview with steps from the algorithm). Smoothing begins with filtering SMC which provides the forward weights and particles $\{\mathbf{z}_{1:T}^{1:K},w_{1:T}^{1:K}\}$ (\textit{step 1}). With outputs from filtering SMC, SVO proceeds to generate backward trajectories sequentially. At time $T$, for each trajectory we will draw $M$ \textit{subparticles} from a continuous-domain conditional kernel (\textit{step 3}). While the final time step requires some care, these subparticles will be used to initialize \textit{subweights} relative to the conditional kernel (\textit{step 4}). The subweights in turn, are used to update the corresponding particle by drawing a backward index from a resampling process (\textit{step 5}). The trajectory is initialized with the selected particle and extended sequentially (\textit{step 6}). SVO iterates by drawing $M$ subparticles from a continuous-domain backward proposal for each of the $K$ trajectories at the current time step (\textit{step 9}). SVO then computes subweights for each subparticle (\textit{step 10}) in order to select a single backward particle from the set of $M$ candidates (\textit{step 11}).  Finally the backward kernel is evaluated using the chosen resampled particle (\textit{step 13}). 
 The output of this procedure is a collection of particle trajectories from the smoothing posterior that are used to define a variational objective.
 
\begin{figure*}
\centering
\includegraphics[width=0.9\textwidth]{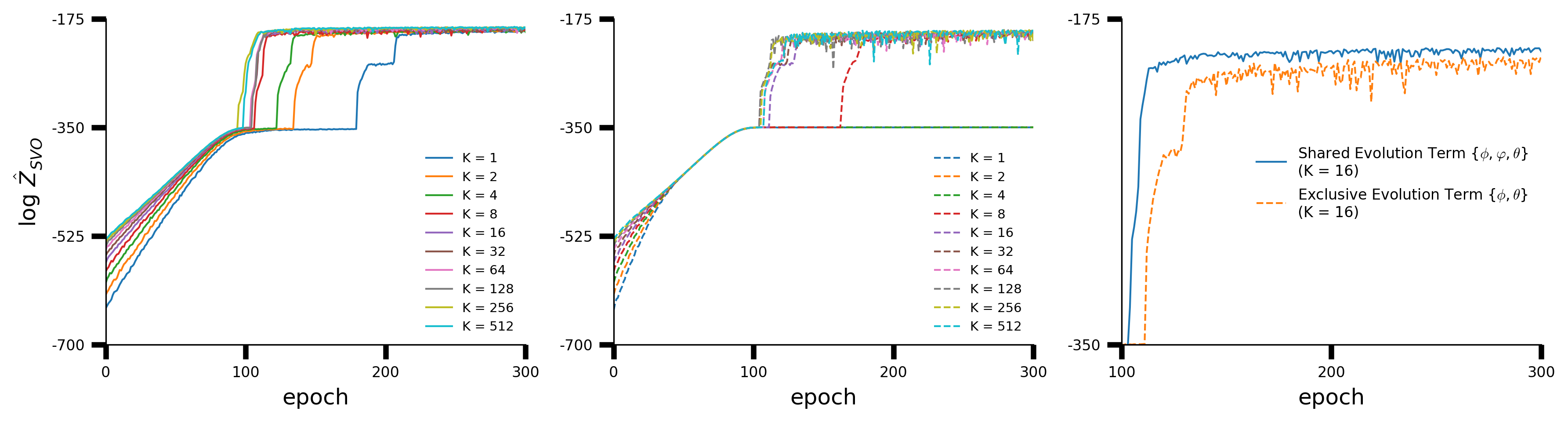}
\caption{ELBO convergence across epochs for SVO using exclusive parameters $\theta, \phi$ and shared parameters $\theta, \varphi, \phi$; (left) log $\hat{\mathcal{Z}}_{SVO}$ across epochs as $K$ increases using shared evolution network; (center)  log $\hat{\mathcal{Z}}_{SVO}$ across epochs as $K$ increases using independent evolution networks; (right) log $\hat{\mathcal{Z}}_{SVO}$ convergence for shared vs independent evolution networks with $K=16$ highlighting faster convergence to a higher ELBO.}
\label{fig:log_ZSMC}
\end{figure*}

\paragraph{Objective Function}
We introduce a \textit{continuous} reverse-dynamics proposal $q (\mathbf{z}_t | \mathbf{z}_{t+1}, \mathbf{x}_{1:T})$ that is used to sample $M$ subparticles for each $k \in \{1,\cdots,K\}$, $\tilde{\mathbf{z}}_t^{k,1:M} \sim q (\mathbf{z}_t | \tilde{\mathbf{z}}_{t+1}^k, \mathbf{x}_{1:T})$. 
These samples are used 
to define subweights as follows
\begin{align}
    \omega_{t|T}^{k,m} &\propto \nonumber \\
    &\int p(\mathbf{z}_{t-1}, \tilde{\mathbf{z}}_t^{k,m}|\mathbf{x}_{1:t-1}) d\mathbf{z}_{t-1} \frac{f(\tilde{\mathbf{z}}_{t+1}^k | \tilde{\mathbf{z}}_t^{k,m}) g(\mathbf{x}_t| \tilde{\mathbf{z}}_t^{k,m})}{q(\tilde{\mathbf{z}}_t^{k,m}| \tilde{\mathbf{z}}_{t+1}^k, \mathbf{x}_{1:T})} \nonumber\\
    &\approx
    \left[ \sum\limits_{j=1}^{K} w_{t-1}^j f(\tilde{\mathbf{z}}_t^{k,m}|\mathbf{z}_{t-1}^j) \right] \frac{f( \tilde{\mathbf{z}}_{t+1}^k | \tilde{\mathbf{z}}_t^{k,m})g(\mathbf{x}_t| \tilde{\mathbf{z}}_t^{k,m})}{q(\tilde{\mathbf{z}}_t^{k,m}| \tilde{\mathbf{z}}_{t+1}^k, \mathbf{x}_{1:T})}.
\end{align}
A single particle is selected by sampling an index proportional to the subweight $\omega_{t|T}$:  $b_t^k\sim\textsc{Categorical}(b_t^k|\omega_{t|T}^{k,1},\cdots,\omega_{t|T}^{k,M})$,  $\tilde{\mathbf{z}}_t^k \leftarrow \tilde{\mathbf{z}}_t^{k, b_t^k}$.
This modified particulate distribution now generates hidden states from a \textit{continuous} domain given the future state and all observations. Repeating this process sequentially in the reverse-time direction 
produces $K$ i.i.d. sample trajectories, $\{\tilde{\mathbf{z}}_{1:T}^{1:K}\}$ (see Algorithm ~\ref{alg:svoFFBSi}). 
\begin{algorithm2e}
\SetAlgoNoLine
\caption{Particle Smoothing Variational Objectives}
\begin{enumerate}
    \item[1.] Perform forward filtering to obtain $\{\mathbf{z}_{1:T}^{1:K},w_{1:T}^{1:K}\}$
    \item[2.] Initialization. For $k = 1, \cdots, K:$
    \begin{enumerate}
        \item[3.] Sample $M$ subparticles: 
        $\{ \tilde{\mathbf{z}}_T^{k,m} \}_{m=1}^M \sim q (\cdot | \mathbf{x}_{1:T})$
        \item[4.] Initialize subweight for each subparticle:
        \begin{align*}
            \omega_{T|T}^{k,m} \propto  \Big [\sum_j w_{T-1}^j f(\tilde{\mathbf{z}}_T^{k,m}|\mathbf{z}_{T-1}^j) \Big]
             \frac{g(\mathbf{x}_T|\tilde{\mathbf{z}}_T^{m,k})}{q(\tilde{\mathbf{z}}_T^{k,m} | \mathbf{x}_{1:T})}
        \end{align*}
        \item[5.] Sample index: $ b_T^k \sim \textsc{Categorical}(\cdot |\omega_{T|T}^{k,1}, \cdots,  \omega_{T|T}^{k,M})$
    \item[6.] 
    Set backward particle: $ \tilde{\mathbf{z}}_T^k \leftarrow \tilde{\mathbf{z}}_T^{k, b_t^k} $, $\omega_{T|T}^k\leftarrow \omega_{T|T}^{k, b_t^k}$
    
    \item[7.] Evaluate the backward proposal:
    $\Omega_T^k \coloneqq M\cdot\omega_{T|T}^{k}\cdot 
    q (\tilde{\mathbf{z}}_T^{k} | \mathbf{x}_{1:T})$  
    \end{enumerate}
    
    \item[8.] Backward Simulation. For $t= T-1, \cdots, 1$ and $k = 1,\cdots, K:$ 
    \begin{enumerate}
        \item[9.] Sample $M$ subparticles from reverse-dynamics proposal: 
        
        $ \{ \tilde{\mathbf{z}}_{t}^{k, m} \}_{m=1}^M \sim q(\cdot|\tilde{\mathbf{z}}_{t+1}^k, \mathbf{x}_{1:T}) $
            \item[10.] Compute subweights:
                
     $\omega_{t|T}^{k,m}  \propto$  $\sum\limits_{j}^{} w_{t-1}^j f(\tilde{\mathbf{z}}_t^{k,m}| \mathbf{z}_{t-1}^j)  \times \frac{f(\tilde{\mathbf{z}}_{t+1}^k|\tilde{\mathbf{z}}_t^{k,m})g(\mathbf{x}_t|\tilde{\mathbf{z}}_t^{k,m})}{q(\tilde{\mathbf{z}}_t^{k,m}|\tilde{\mathbf{z}}_{t+1}^k, \mathbf{x}_{1:T})}$
            \item[11.] Sample index: $ b_t^k \sim \textsc{Categorical}(\cdot| \omega_{t|T}^{k,1}, \cdots, \omega_{t|T}^{k,M}) $
    \item[12.] Set backward particle: $\tilde{\mathbf{z}}_t^k \leftarrow \tilde{\mathbf{z}}_t^{k,b_t^k}$, $\omega_{t|T}^k\leftarrow \omega_{t|T}^{k, b_t^k}$
    \item[13.] Evaluate the backward proposal:
    $\Omega_t^k = M\cdot\omega_{t|T}^{k}\cdot 
    q (\tilde{\mathbf{z}}_t^{k} |\tilde{\mathbf{z}}_{t+1}^{k},\mathbf{x}_{1:T})$
    
    \end{enumerate}
\item[14.]\Return 
 \begin{equation*}
     \hat{\mathcal{L}}_{SVO}^K(\tilde{\mathbf{z}}_{1:T}^{1:K},\mathbf{x}_{1:T}) = \log \left(\frac{1}{K}\sum_{k=1}^K \frac{p(\tilde{\mathbf{z}}_{1:T}^{k},\mathbf{x}_{1:T} )}{\prod_{t=1}^T \Omega_t^k}\right)
 \end{equation*}
 \end{enumerate}
 \label{alg:svoFFBSi}
\end{algorithm2e}
The approximate posterior and variational objective are defined below via Algorithm~\ref{alg:svoFFBSi} (for a detailed treatment see Appendix C):
\begin{equation}
\mathcal{L}_{SVO} \coloneqq \underset{q}{\mathbb{E}}\left[\log \hat{\mathcal{Z}}_{SVO} \right]\, ,
\  \ \hat{\mathcal{Z}}_{SVO} \coloneqq \frac{1}{K} \sum_{k=1}^K \frac{p(\tilde{ \mathbf{z}}_{1:T}^{k},\mathbf{x}_{1:T})}{q(\tilde{ \mathbf{z}}_{1:T}^{k}|\mathbf{x}_{1:T})}\, ,
\end{equation}
where $\quad q(\tilde{\mathbf{z}}_{1:T}^{k} | \mathbf{x}_{1:T}) \coloneqq$
\begin{align}
M^T  \omega_{T|T}^{k}  
q(\tilde{\mathbf{z}}_T^{k} | \mathbf{x}_{1:T})\prod_{t=1}^{T-1} \left[ \omega_{t|T}^{k}  
q(\tilde{\mathbf{z}}_t^{k}| \tilde{\mathbf{z}}_{t+1}^{k}, \mathbf{x}_{1:T}) \right]. 
\end{align}
We note that while the sequence of target distributions is filtered, our objective is constructed using samples from a smoothing posterior.  This heuristic facilitates smoothing the target when performing VI to simultaneously train $p(\mathbf{Z}|\mathbf{X})$ and $q(\mathbf{Z}|\mathbf{X})$ by pulling $p(\mathbf{Z}|\mathbf{X}) \rightarrow q(\mathbf{Z}|\mathbf{X})$. This functional dependence motivates sharing the transition function between proposal and target.
\begin{theorem}
$\hat{Z}_{SVO}$ is an unbiased estimate of $p(\mathbf{x}_{1:T})$.
\end{theorem}\begin{table*}
\begin{prop}
Assume that the first four moments of $w_t^{1}$ and $\nabla w_t^{1}$ are all finite and their variances are non-zero for $t \in 1:T$, then the signal-to-noise ratio converges at the following rate: 
\begin{equation}
\begin{aligned}
\text{SNR}_K(\theta, \varphi, \phi) 
= \left|\frac{\nabla \log Z  +  \sum_{t=2}^T \sum_{t'\geq t+1}^T \mathbb{E}\left[ \nabla \frac{w_{t-1}^1}{ Z_{t-1} } \cdot \frac{(w_{t'}^1 -Z_{t'})^2}{2Z_{t'}^2} \Big|\left[a_{t-1}^1=1\right]\right] + \mathcal{O}(\nicefrac{1}{K})}
{ \sqrt{ \nicefrac{1}{K} \left\{ \sum\limits_{t=1}^T \mathbb{E} \left[( \nabla \frac{w_t^1}{Z_t})^2\right]  + \sum\limits_{t'\neq t, t'=1}^T \sum \limits_{t=1}^T \sqrt{ \Var \left[ \nabla \frac{w_t^1}{Z_t}\right] \Var \left[\nabla \frac{w_{t'}^1}{Z_{t'}}\right]} \right\} + \mathcal{O}(\nicefrac{T^2}{K^2}) } } \right|
\end{aligned}
\end{equation}
\end{prop}
where $Z = p_\theta (\mathbf{x}_{1:T})$ and $Z_t = p_\theta (\mathbf{x}_t | \mathbf{x}_{1:t-1})$ for $t \in \{1, \cdots, T \}$. Further assuming the resampling bias $\sum_{t=2}^T \sum_{t'\geq t+1}^T \mathbb{E}\left[ \nabla \frac{w_{t-1}^1}{ Z_{t-1} } \cdot \frac{(w_{t'}^1 -Z_{t'})^2}{2Z_{t'}^2} \Big|\left[a_{t-1}^1=1\right]\right] = \mathcal{O}(1)$ leads to $\text{SNR}_K (\theta, \phi, \varphi) = \mathcal{O}(\sqrt{K})$.
\begin{proof}
See Appendix D.
\end{proof}
\end{table*}
\begin{proof}
We give an intuitive sketch of the result here and provide a formal proof in Appendix C. Since the expectation is a linear operator, we consider the simple case of $K=1$. Here we make a slight notation change where we relabel the selected particles as $\tilde{\mathbf{z}}_{1:T}^1$,and the unselected ones as  $\tilde{\mathbf{z}}_{1:T}^{2:M}$. The generating process of all subparticles is described by the proposal distribution:  
\begin{equation}
    Q(\tilde{\mathbf{z}}_{1:T}^{1:M}) = \Omega_T \cdot q(\tilde{\mathbf{z}}_T^{2:M}|\mathbf{x}_{1:T}) \prod_{t=1}^{T-1} \Omega_t \cdot q(\tilde{\mathbf{z}}_t^{2:M} | \tilde{\mathbf{z}}_{t+1}^1, \mathbf{x}_{1:T}).
\end{equation} We extend our target into the whole particulate space:
\begin{equation}
    P(\mathbf{x}_{1:T}, \tilde{\mathbf{z}}_{1:T}^{1:M}) = p(\mathbf{x}_{1:T}, \tilde{\mathbf{z}}_{1:T}^1) r (\tilde{\mathbf{z}}_{1:T}^{2:M} | \tilde{\mathbf{z}}_{1:T}^{1}, \mathbf{x}_{1:T}).
\end{equation}
 A convenient choice of 
 \begin{align}
     r(\tilde{\mathbf{z}}_{1:T}^{2:M}&|\tilde{\mathbf{z}}_{1:T}^{1},\mathbf{x}_{1:T}) \\
     &=  q(\tilde{\mathbf{z}}_T^{2:M} | \tilde{\mathbf{z}}_{T}^1, \mathbf{x}_{1:T}) \prod_{t=1}^{T-1} q(\tilde{\mathbf{z}}_t^{2:M} |\tilde{\mathbf{z}}_{t+1}^1,\mathbf{x}_{1:T}) \nonumber
 \end{align}would lead to cancellation of terms in the target and the proposal, recovering the desired $\hat{Z}_{SVO}$. The unbiasedness follows naturally. 
\end{proof}

\paragraph{Parameterizing the Filtering and Smoothing Proposals}
In the forward filtering pass, we define the proposal distribution as follows:
\begin{align}
q_{\phi,\varphi}&(\mathbf{z}_{1:T}^{k}|\mathbf{x}_{1:T}) 
\propto 
 \underbrace{f_{\varphi}(\mathbf{z}_1^{k})}_{initial\ state} \prod\limits_{t=1}^{T} \underbrace{h_{\phi}(\mathbf{z}_t^{k}|\mathbf{x}_{t})}_{encoding}
 \\
 & 
 \prod\limits_{t=2}^{T}
 \underbrace{\textsc{Categorical}(a_{t-1}^{k}|\bar{w}_{t-1}^{1:K})}_{resampling} 
 \underbrace{f_{\varphi}(\mathbf{z}_t^{k}|\mathbf{z}_{t-1}^{a_{t-1}^{k}})}_{transition}\ , \nonumber
 \label{eq:svo}
\end{align}
where the proposal density factorizes into evolution and encoding functions,
\begin{align}
f_{\varphi}(\mathbf{z}_t|\mathbf{z}_{t-1}) = \mathcal{N}(\psi(\mathbf{z}_{t-1}), \Sigma), \ \
h_{\phi}(\mathbf{z}_t|\mathbf{x}_t) = \mathcal{N}(\gamma(\mathbf{x}_t), \Lambda).
\end{align}
We define $\psi: \mathbb{R}^{d_\mathbf{z}} \rightarrow \mathbb{R}^{d_\mathbf{z}}$ and $\gamma: \mathbb{R}^{d_\mathbf{x}} \rightarrow \mathbb{R}^{d_\mathbf{z}}$ as nonlinear time invariant functions represented with deep neural networks. The covariances $\Sigma$ and $\Lambda$ are taken as time invariant trainable parameters or nonlinear functions of the latent space. 
This proposal choice allows the transition term of the inference network $f_{\varphi}(\mathbf{z}_t|\mathbf{z}_{t-1})$ to share the parameters $\varphi$ defining $\{\psi, \Sigma\}$ with the transition term $f_{\varphi}(\mathbf{z}_t|\mathbf{z}_{t-1})$ of the target defined in Eq. (\ref{eq:mgen}) ~\cite{anh2018autoencoding,NIPS2017_7235,pmlr-v84-naesseth18a}. The evolution term of the variational posterior is exact, retaining both tractability and expressiveness. 

The transition and emission densities are specified as follows:
\begin{align}
f_{\varphi}(\mathbf{z}_t|\mathbf{z}_{t-1}) = \mathcal{N}(\psi(\mathbf{z}_{t-1}), \Sigma), \ \ g_\theta(\mathbf{x}_t|\mathbf{z}_t) = \mathcal{N}(\upsilon(\mathbf{z}_t), \Gamma)\ .
\end{align}
The decoding term is defined using a deterministic nonlinear rate function $\upsilon:  \mathbb{R}^{d_\mathbf{z}} \rightarrow \mathbb{R}^{d_\mathbf{x}}$ represented with a deep network and a noise model that need not be conjugate. Without loss of generality we consider a Gaussian emission density. The backward proposal defining the smoothing distribution below
\begin{equation}
q (\mathbf{z}_t | \mathbf{z}_{t+1}, \mathbf{x}_{1:T}) \propto r(\mathbf{z}_t|\zeta(\mathbf{z}_{t+1}))e(\mathbf{z}_t|\chi(\mathbf{x}_{1:T})),
\label{backprop}
\end{equation}
is specified using nonlinear time invariant functions $\zeta: \mathbb{R}^{d_\mathbf{z}} \rightarrow \mathbb{R}^{d_\mathbf{z}}$ and $\chi: \mathbb{R}^{d_\mathbf{X}} \rightarrow \mathbb{R}^{d_\mathbf{z}}$ which we take as deep networks.

\section{SNR of Gradient Estimators in Filtering SMC}
\label{secsnr}
$\mathcal{L}_{\text{SMC}}$ is a consistent estimator of the log marginal likelihood under some mild conditions \cite{NIPS2017_7235}. Intuition suggests increasing the number of particles $K$ provides a better surrogate objective. However, \cite{rainforth2018tighter} points out that the SNR of the inference network gradient estimator decreases to 0 as $K$ increases in the IWAE setting. \cite{anh2018autoencoding} extends the result to the filtering SMC without providing theoretical evidence. Here, we argue that the result does not generalize to SMC due to the resampling step. 
 Formally, for a gradient estimator of $\mathcal{L}_{\text{SMC}}$ (denoted $\Delta_K$), constructed by $K$ particles, the SNR is defined as:
\begin{equation}
\text{SNR}_K  = \Bigg|\frac{\mathbb{E} [\Delta_K] }{\sqrt{\Var [\Delta_K]}}  \Bigg|.
\end{equation}
\begin{figure*}
\centering
\includegraphics[width=0.9\textwidth]{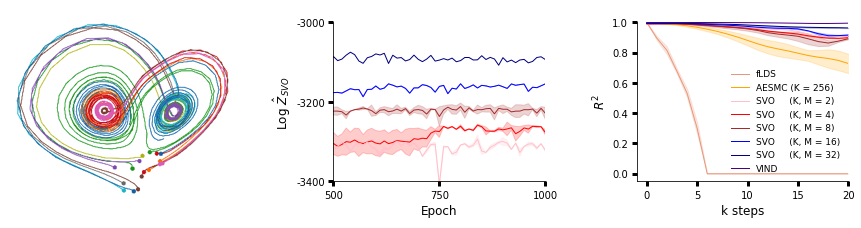}
\caption{Summary of the Lorenz results: (left) latent trajectories inferred from nonlinear 10D observations; (center) $\log \hat{Z}_{SVO}$ as $K$, $M$ increase (legend on the right). Larger $K,M$ produce higher ELBO values; (right) $R^2_{k}$ on the dimensionality reduction task illustrating near-perfect reconstruction at 20 steps ahead on the validation set. Results averaged over 3 random seeds.}
\label{lorenz}
\end{figure*}
For the SNR of $\nabla \mathcal{L}_K$, we have the following Proposition 1. 
We add empirical evidence to this result in Section 6. We consider three types of stochastic gradient estimators. A full definition is given in the Appendix.
\begin{enumerate}[leftmargin=2\parindent]
    \item The biased estimator without resampling gradient, $\nabla \mathcal{L}_K$.
    \item The unbiased estimator, $\nabla \mathcal{L}_K + \textsc{Categorical}$.
    \item The relaxed estimator,  $\nabla \mathcal{L}_K + \textsc{Concrete}(\lambda)$ 
     \cite{jang2016categorical,maddison2016concrete}.
\end{enumerate}

\nocite{doi:10.1089/cmb.2018.0176}
\section{Related Work}

AESMC~\cite{anh2018autoencoding}, FIVO~\cite{NIPS2017_7235} and VSMC~\cite{pmlr-v84-naesseth18a} are three closely related methods that construct a variational objective by performing filtering SMC to estimate the log marginal likelihood. 
AESMC, FIVO and VSMC define expressive variational families in both generative and recognition models, however without using future observations to infer the current latent state they may fail to capture long-term dependencies. VSMC in particular draws a single sample at the final time step to produce a
trajectory from the corresponding ancestral path. While this heuristic produces \textit{one} sample conditioned on all observations, the resulting path is not used to construct the surrogate ELBO which is filtered. 

It is important to utilize information from the complete observation sequence to approximate the current latent state. Two smoothing methods for inference in non-conjugate SSMs are GfLDS~\cite{archergao,archer} and VIND~\cite{vind2}.
GfLDS is a generative model and approximation for linear latent dynamics together with nonlinear emission densities. Building upon this, VIND 
is governed by nonlinear latent dynamics and emissions. 
GfLDS and VIND both require inverting a block-tridiagonal matrix which mixes components of state space through the inverse covariance. This incurs a complexity of $\mathcal{O}(Td_{\mathbf{z}}^3)$ where $T$ is the length of the time series and $d_{\mathbf{z}}$ is the state dimension. 
In contrast, SVO can perform smoothing in $\mathcal{O}(TK^2d_{\mathbf{z}})$ operations. 
An alternative approach is to directly modify the target distribution in SMC to achieve smoothing. 
TVSMC~\cite{lawsontwisted} and SMC-Twist~\cite{lindsten2018graphical} augment the intermediate target distribution with a twisting function, which in turn is approximated with deterministic algorithms such as temporal difference learning and Laplace approximation. 
When applied to nonlinear time series it was reported that TVSMC underperforms relative to filtering using VSMC~\cite{lawsontwisted}. 

\begin{figure*}
\centering
\includegraphics[width=0.925\textwidth]{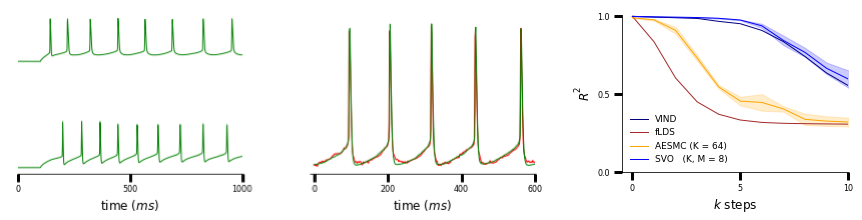}
\caption{Summary of the Allen results: (left) two trials from the dataset; (center) the data against the predicted observation value using the dynamics learned over a rolling window ten steps ahead on the validation set. Hyperpolarization and depolarization nonlinearities are predicted by the inferred dynamics; 
(right) $R^2_k$ with $K,M=8$ particles. SVO outperforms GfLDS and AESMC with $K=64$. Results are averaged across 3 random seeds.}
\label{allen}
\end{figure*}

\section{Experimental Results}\label{results}
In order to quantify the performance of the trained dynamics, we compute the $k$-step mean squared error (MSE) and its normalized version, the $R^2_k$. To do so, the trained transition function is applied to the latent state without any input data over a rolling window of $k$ steps into the future. The emission function is then used to obtain a prediction  
$\hat{\mathbf{x}}_{t+k}$ which we compare with the observation $\mathbf{x}_{t+k}$. 
\begin{align}
\text{MSE}_k &=  \sum_{t=1}^{T-k} \left( \mathbf{x}_{t+k} - \hat{\mathbf{x}}_{t+k} \right)^2 \\
R^2_k &= 1 - \frac{\text{MSE}_k}{\sum_{t=1}^{T-k} \left( \mathbf{x}_{t+k} - \bar{\mathbf{x}}_k \right)^2} \,, \nonumber
\end{align}
where $\bar{\mathbf{x}}_k$ is the average of $\mathbf{x}_{k+1:T}$.
We note that the ELBO is not a performance statistic that generalizes across models. In contrast, the $R^2_k$ provides a metric to quantify the inferred dynamics. This procedure is defined in~\cite{vind2}.

\paragraph{Fitzhugh-Nagumo}
The Fitzhugh-Nagumo (FN) system is a two dimensional simplification of the Hodgkin-Huxley model. The FN provides a geometric interpretation of the dynamics of spiking neurons and is described by two independent variables $V_t$ and $W_t$ with cubic and linear functions, 
\begin{align}
\dot V &= V - V^3/3 - W + I_{ext} \,\label{fneq} \\ \dot W &= a(bV - cW). \nonumber
\end{align}
Eq. (\ref{fneq}) was integrated over 200 time points with $I_{ext} = 1$ held constant and $a = 0.7, b = 0.8, c = 0.08$. The initial state was sampled uniformly over $[-3,3]^2$ to generate 100 trials using 66 for training, 17 for validation and 17 for testing. We emphasize that dimensionality expansion is intrinsically harder than dimensionality reduction due to a loss of information.
A one-dimensional Gaussian observation is defined on $V_t$ with $\mathbf{x}_t = \mathcal{N}(V_t,0.01)$. SVO is used to recover the two dimensional phase space and latent trajectories $\mathbf{z}_t = (V_t, W_t)$ of the original system. This task requires using information from future observations to correctly infer the initial state. 
Fig. \ref{fhn} shows the results of the FN experiment. The left panel displays the original system. The center panel displays the learned dynamics and inferred trajectories on the test set using SVO to perform dimensionality expansion. Initial points (denoted with markers) located both inside and outside of the limit cycle in the original system are topologically invariant in the reconstruction.
The right panel shows the $R^2_k$ comparison across models. AESMC with $K=1024$ gives an $R^2_{30}=0.954$ in contrast to SVO with $K=32, M=32$ which gives an $R^2_{30} = 0.993$. SVO outperforms AESMC and GfLDS. 

\paragraph{Sharing Transition Terms}
We study the effect of sharing the transition function between the proposal and target distribution. Fig. \ref{fig:log_ZSMC} illustrates the ELBO convergence as the number of particles $K$ is increased. The left panel plots ELBO for SVO with network parameters 
shared between proposal and target. Increasing $K$ produces a faster convergence and lower stochastic gradient noise. The center panel illustrates separate evolution networks for the proposal and the target. In contrast to sharing the transition function, separate evolution networks require a larger number of epochs for corresponding value of $K$. The ELBO obtains a lower value with larger stochastic gradient noise. The right panel juxtaposes shared and separate transition functions for $K=16$ particles.

\paragraph{Lorenz Attractor} The Lorenz attractor is a chaotic nonlinear dynamical system defined by 3 independent variables,
\begin{align}
  \dot{z}_1 &= \sigma(z_2-z_1) \,, \nonumber \\ 
  \dot{z}_2 &= z_1(\rho - z_3) - z_2 \,, \label{eq:chaos} \\
  \dot{z}_3 &= z_1z_2 - \beta z_3\, \nonumber .
\end{align}
Eq. (\ref{eq:chaos}) is integrated over 250 time points with $\sigma=10, \rho=28, \beta=8/3$ by generating randomized initial states in $[-10,10]^3$. A $\mathbf{z}$-dependent neural network is used to produce ten dimensional nonlinear Gaussian observations with 100 trials, 66 for training, 17 for validation and 17 for testing. Fig. \ref{lorenz} provides the results of the Lorenz experiment. 
The left panel provides the inferred latent paths illustrating the attractor. The center plot provides $\log \hat{Z}_{SVO}$ as $K$, $M$ increase (legend on the right). Larger $K,M$ produce higher ELBO values. 
The right panel displays the $R^2_k$ comparison with $d_\mathbf{z}=3$. Results are averaged over 3 random seeds. Increasing $K, M$ produces $R^2_k$ improvements. SVO with $K, M = 2$ gives a higher $R^2_k$ than both GfLDS and AESMC using $K=256$.

\paragraph{Electrophysiology Data}
Neuronal electrophysiology data was downloaded from the Allen Brain Atlas~\cite{allennature}. Intracellular voltage recordings from primary Visual Cortex of mouse, area layer 4 were collected. A step-function input current with an amplitude between 80 and 151pA was applied to each cell.  A total of 40 trials from 5 different cells were split into 30 trials for training and 10 for validation. Each trial was divided into five parts and down-sampled from 10,000 time bins to 1,000 time bins in equal intervals. Each trial was normalized by its maximal value.
Fig.~\ref{allen} summarizes the  Allen experiment. The left panel provides two trials of the 1D observations from the training set. 
The center panel illustrates the predicted observation using the dynamics learned over a rolling window ten steps ahead on the validation set. SVO captures hyperpolarization and depolarization nonlinearities when appying the inferred dynamics. The right panel displays the $R^2_k$ comparison with $d_\mathbf{z}=3$. SVO outperforms AESMC and GfLDS.

\begin{figure*}[t!]
\centering
\includegraphics[width=0.95\textwidth]{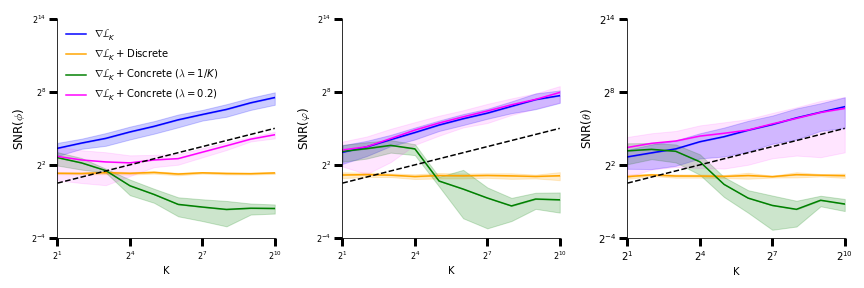}
\caption{Convergence of SNRs of gradient estimators in the encoder network (left), transition network (center) and decoder network (right) with increasing $K$. Distinct solid lines correspond to empirical SNRs of the four gradient estimators, averaging over 6 random seeds. The black dashed line with slope 1 illustrates a signal-to-noise-ratio of convergence rate $\mathcal{O}(\sqrt{K})$.}
\label{snr}
\end{figure*}

\paragraph{SNR Gradient Estimators}
We report the $l_2$ norm of empirical SNRs for the encoder network $(\phi)$, evolution network $(\varphi)$ and decoder network $(\theta)$, where the gradient is taken with respect to $\phi$, $\varphi$ and $\theta$ correspondingly. Fig.~\ref{snr} presents four gradient estimators where the expectation and variance are calculated using $N=100$ gradient samples collected in the middle training stage of running filtering SMC on Fitzhugh-Nagumo data.
The gradient estimator that ignores the resampling step possesses an SNR of convergence rate $\mathcal{O}(\sqrt{K})$, which aligns with the theoretical result. Similarly this holds for the relaxed $\textsc{Concrete}$ gradient estimator with a constant temperature ($\lambda=0.2$). 
The unbiased \textsc{Categorical} resampling gradient and the relaxed $\textsc{Concrete}$ gradient with  decreasing temperature ($\lambda = K^{-1}$) suffer from large variance, leading to a relatively low and even vanishing SNR for increasing $K$. Moreover, the level of relaxation $\lambda$ in the $\textsc{Concrete}$ gradient estimator leads to different behaviors of SNR. These observations imply that introducing bias reduces the variance and mitigates the degradation of the SNR with increasing $K$.

\paragraph{Conclusion}
We have introduced SVO, a novel algorithm and approximate posterior constructed from recursive backward sampling. SMC's resampling step introduces challenges for standard reparameterization due to the \textsc{Categorical} distribution. We address this by presenting theoretical and empirical evidence to support the choice of biased gradient estimates. SVO consistently outperforms filtered objectives across  nonlinear dynamical systems.  SVO is written in TensorFlow. An implementation is publicly  available online. 

\bibliography{psvo}
\bibliographystyle{aaai}

\clearpage

\twocolumn[\section{ Supplementary Material\\ \text{}}]

\subsection*{Appendix A. Gradient estimators}

\begin{enumerate}
    \item The biased gradient estimator without resampling $\nabla \mathcal{L}_K$ is implemented by simply taking the gradient of the estimated variational objective, $\log \hat{Z}_{\text{SMC}}$.
    \item The unbiased gradient estimator,
    \begin{align*}
        &\nabla \mathcal{L}_K + \textsc{Categorical} =\\ &\nabla \mathcal{L}_K + \nabla \log \prod_{t=1}^{T-1} \prod_{k=1}^K \textsc{Categorical} (a_t^{k} | w_t^{1:K}) \cdot \mathcal{L}_K. \nonumber
    \end{align*} 
    The derivation can be found in \cite{anh2018autoencoding} or \cite{NIPS2017_7235} .
    \item For the relaxed gradient estimator,
    \begin{align*}
        &\nabla \mathcal{L}_K + \textsc{Concrete}(\lambda) =\\ 
        &\nabla \mathcal{L}_K + \nabla \log \prod_{t=1}^{T-1} \prod_{k=1}^K \textsc{Concrete} (a_t^{k} | w_t^{1:K},\lambda) \cdot \mathcal{L}_K. \nonumber
    \end{align*}
    We use the $\textsc{Concrete}$ distribution to resample particles, and then directly evaluate the gradient of the objective.

\end{enumerate}


\subsection*{Appendix B. Nonlinear Evolution of the Covariance Matrix} 
In the Allen experiment we define a locally linear covariance matrix to express nonlinear $\mathbf{z}$-dependence on the latent state. This permits the transition term in the latent state to depend on a nonlinear noise parameter. 
 To ensure smoothness of the latent trajectories, the difference between the covariance matrix and a constant matrix $\mathbb{C}$ should be small,
\begin{equation*}
\max_{} |\mathbf{Q}(\mathbf{z}_t) - \mathbb{C}| \lesssim 0.1\, .
\end{equation*}
To achieve this, the covariance matrix is parameterized by a constant plus a scalar times a symmetric matrix whose components are a nonlinear function of the latent state,
\begin{equation*}
\mathbf{Q}(\mathbf{z}_t) =  \mathbb{C} + \alpha \cdot \mathbf{\Sigma}(\mathbf{z}_t)\, .
\end{equation*}
We observe improvements in the $R^2_k$ on the Allen task when incorporating a locally linear covariance matrix. On this task $\alpha = 1e-1$, $\mathbb{C} = \mathbb{I}\cdot\sigma^2$ where $\sigma^2$ is a trainable variable and the components of $\mathbf{\Sigma}(\mathbf{z}_t)$ are taken as the output of a two-layer feed forward network. 
At different positions in phase space the system can suppress or enhance its sensitivity to noise.  

\subsection{Appendix C. Proof of Theorem 1.}

\paragraph{Theorem 1.}
$\hat{Z}_{SVO}$ is an unbiased estimate of $p(\mathbf{x}_{1:T})$.

\begin{equation*}
    \mathbb{E}_{\hat{\mathbf{z}}_{1:T}^{1:K, 1:M}} \left[\frac{1}{K} \sum_{k=1}^K \frac{p(\hat{\mathbf{z}}_{1:T}^{k}, \mathbf{x}_{1:T})}{\prod_{t=1}^T \Omega_t^K}  \right] = p(\mathbf{x}_{1:T})
\end{equation*}
\begin{proof}
We will define auxiliary variables $\lambda$ and distributions $q(\lambda|x)$, $q(z|\lambda,x)$, and $r(\lambda|z,x)$ such that
\begin{equation*}
    \hat{\mathcal{Z}}_{SVO} \equiv \hat{p}(x) =  \frac{p(x,z)r(\lambda|z,x)}{q(z,\lambda|x)} = \frac{p(x,z)r(\lambda|z,x)}{q(z|\lambda,x)q(\lambda|x)},
\end{equation*}
where $z,\lambda\sim q(z,\lambda|x)$. For a treatment of auxiliary random variables see~\cite{NIPS2018_7699,lawsonaux}.
Here the auxiliary latent variables are the unselected subparticles,
\begin{align*}
    \lambda = \{ \tilde{\mathbf{z}}_{1:T}^{\neg b_{1:T}^{1:K}}\}.
\end{align*}
For convenience, we omit the conditioning on the forward system. To further simplify notation, we will rearrange particles to omit the backward ancestor indices by defining $\hat{\mathbf{z}}_t^{k,1} \leftarrow \tilde{\mathbf{z}}_t^{k,b_t^k}$, $\hat{\omega}_{t|T}^k\leftarrow \omega_{t|T}^{k, b_t^k}$ and $\hat{\mathbf{z}}_t^{k, 2:M} \leftarrow \tilde{\mathbf{z}}_t^{k, \neg b_t^k}, \hat{\omega}_{t|T}^{k, 2:M} \leftarrow \omega _{t|T}^{k, \neg b_t^k}$. By the linearity of expectation, it suffices to show the case of $K=1$ (as a result, for clarity, we will omit $k,$ in the superscripts):
\begin{equation*}
    \mathbb{E}_{\hat{\mathbf{z}}_{1:T}^{1:M}} \left[ \frac{p(\hat{\mathbf{z}}_{1:T}^{1}, \mathbf{x}_{1:T})}{\prod_{t=1}^T \Omega_t}  \right] = p(\mathbf{x}_{1:T})
\end{equation*}
We begin by expressing the sampling distribution (the true generating probability) of $\hat{\mathbf{z}}_{1:T}^{1:M}$ as factorizing:
\begin{equation*}
 Q(\hat{\mathbf{z}}_{1:T}^{1:M} | \mathbf{x}_{1:T}) = Q(\hat{\mathbf{z}}_T^{1:M} | \mathbf{x}_{1:T}) \prod_{t=1}^{T-1} Q( \hat{\mathbf{z}}_{t}^{1:M} | \hat{\mathbf{z}}_{t+1}^1, \mathbf{x}_{1:T}).
\end{equation*}
Consider the last time step, 
\begin{align*}
&Pr(\hat{\mathbf{z}}_T^1=z_1, \hat{\mathbf{z}}_T^{2:M}=z_{2:M}) \\
&= \left[\prod_{m=1}^M q(z_m | \mathbf{x}_{1:T})\right] \cdot  \sum_{i=1}^M Pr(b_T=i | \tilde{\mathbf{z}}_T^i=z_1, \tilde{\mathbf{z}}_T^{\neg i}=z_{2:M}) \\
&= \left[\prod_{m=1}^M q(z_m | \mathbf{x}_{1:T})\right] \cdot  \sum_{b_T=1}^M \frac{\tilde{\omega}^{b_T}}{\tilde{\omega}^{b_T} + \sum_{i \in \neg b_T} \tilde{\omega}^i} \\
&= \left[\prod_{m=1}^M q(z_m | \mathbf{x}_{1:T})\right] \cdot M \cdot \frac{\hat{\omega}^1}{\sum_{i=1}^M \hat{\omega}^m}.
\end{align*}
%
%
Hence, 
\begin{equation*}
    Q(\hat{\mathbf{z}}_T^{1:M} | \mathbf{x}_{1:T}) = \left[\prod_{m=1}^M q(\hat{\mathbf{z}}_T^m | \mathbf{x}_{1:T}) \right] \cdot M \cdot \frac{\hat{\omega}_{T|T}^1}{\sum_{m=1}^M \hat{\omega}_{T|T}^m}.
\end{equation*}
Similarly, we have the following for $t=1, \dots, T-1$, 
\begin{equation*}
    Q(\hat{\mathbf{z}}_t^{1:M}| \hat{\mathbf{z}}_{t+1}^1, \mathbf{x}_{1:T}) = \left[\prod_{m=1}^M q(\hat{\mathbf{z}}_t^m | \hat{\mathbf{z}}_{t+1}^1, \mathbf{x}_{1:T}) \right] M \frac{\hat{\omega}_{t|T}^1}{\sum_{m=1}^M \hat{\omega}_{t|T}^m}.
\end{equation*}
Therefore, 
\begin{equation*}
Q(\hat{\mathbf{z}}_{1:T}^{1:M} | \mathbf{x}_{1:T}) = \underbrace{\left[\prod_{t=1}^T \Omega_t \right]}_{q(z|\lambda,x)} \cdot \underbrace{\prod_{m=2}^M  \left[
q(\hat{\mathbf{z}}_T^m | \mathbf{x}_{1:T}) \prod_{t=1}^{T-1} q(\hat{\mathbf{z}}_t^m | \hat{\mathbf{z}}_{t+1}^1, \mathbf{x}_{1:T}) \right]}_{q(\lambda|x)}.
\end{equation*}
Now, define the target distribution to be:
\begin{align*}
    P(\hat{\mathbf{z}}_{1:T}^{1:M}, \mathbf{x}_{1:T}) &= p(\hat{\mathbf{z}}_{1:T}^1, \mathbf{x}_{1:T})r(\lambda|\mathbf{x}_{1:T},\mathbf{z}_{1:T})\\
    \intertext{where}
    r(\lambda|\mathbf{x}_{1:T},\mathbf{z}_{1:T}^{1:M}) &= q(\lambda|\mathbf{x}_{1:T})\\
    &=\prod_{m=2}^M \left[q(\hat{\mathbf{z}}_T^m | \mathbf{x}_{1:T}) \prod_{t=1}^{T-1}  q(\hat{\mathbf{z}}_t^m | \hat{\mathbf{z}}_{t+1}^1, \mathbf{x}_{1:T})
    \right].
\end{align*}
Then
\begin{align*}
    &\underset{Q}{\mathbb{E}} \left[ \frac{P(\hat{\mathbf{z}}_{1:T}^{1:M}, \mathbf{x}_{1:T})}{Q(\hat{\mathbf{z}}_{1:T}^{1:M}) } \right]
    =\underset{Q}{\mathbb{E}} \left[\frac{p(x,z)r(\lambda|x,z)}{q(z|\lambda,x)q(\lambda|x)} \right]\\
    &=\underset{Q}{\mathbb{E}}\left[ \frac{p(\hat{\mathbf{z}}_{1:T}^1, \mathbf{x}_{1:T}) \prod\limits_{m=2}^M \left[q(\hat{\mathbf{z}}_T^m | \mathbf{x}_{1:T}) \prod\limits_{t=1}^{T-1}  q(\hat{\mathbf{z}}_t^m | \hat{\mathbf{z}}_{t+1}^1, \mathbf{x}_{1:T})\right]}
    {\ \ \ \ \prod\limits_{t=1}^{T} \ \Omega_t \ \times \  \prod\limits_{m=2}^M \left[q( \hat{\mathbf{z}}_T^m | \mathbf{x}_{1:T}) \prod\limits_{t=1}^{T-1} q(\hat{\mathbf{z}}_t^m | \hat{\mathbf{z}}_{t+1}^1,\mathbf{x}_{1:T}) \right]}
    \right] \nonumber\\
    &= \int P(\hat{\mathbf{z}}_{1:T}^{1:M}, \mathbf{x}_{1:T}) d \hat{\mathbf{z}}_{1:T}^{1:M} \nonumber \\
    &= \int p(\hat{\mathbf{z}}_{1:T}^1, \mathbf{x}_{1:T})  \\
    &\times\left[ \int \prod_{m=2}^M
    \left[
    q(\hat{\mathbf{z}}_T^m | \mathbf{x}_{1:T}) \prod_{t=1}^{T-1}  q(\hat{\mathbf{z}}_t^m | \hat{\mathbf{z}}_{t+1}^1, \mathbf{x}_{1:T})
    \right]
    d\hat{\mathbf{z}}_{1:T}^{2:M} \right] d\hat{\mathbf{z}}_{1:T}^1 \nonumber \\
    &= \int p(\hat{\mathbf{z}}_{1:T}^1, \mathbf{x}_{1:T}) d \hat{\mathbf{z}}_{1:T}^1 \\
    &= p(\mathbf{x}_{1:T}).
\end{align*}
\end{proof}

\subsection*{Appendix D. Proof of Proposition 1}

\begin{proof}
It suffices to show the convergence rate of expectation and variance of gradient estimate with respect to $K$.  Throughout the analysis, we will extensively apply the result from \cite{rainforth2018tighter}, and exploit the factorization of the filtering SMC objective: $\hat{Z} \coloneqq \hat{Z}_{\text{SMC}} = \prod_{t=1}^T \hat{Z_t}$ where  $\hat{Z}_t = \frac{1}{K}\sum_{k=1}^K w_t^k$. Assume that $\mathbf{z}_{1:T}^{1:K}$ are obtained by passing the Guassian noise $\epsilon_{1:T}^{1:K}$ through the reparameterization function.
\begin{enumerate}
    \item[1.] Expectation.
\begin{align}
    \mathbb{E} &\left[\nabla \log \hat{Z}\right] = 
        \nabla \mathbb{E} \left[\log \hat{Z}\right] \\ &-  \mathbb{E} \Big[\nabla \log \prod_{t=2}^T \prod_{k=1}^K \textsc{Categorical}(a_{t-1}^k | w_{t-1}^{1:K}) \cdot \log \hat{Z}\Big]\label{expectation} \nonumber
    \end{align}
The expectation decomposes into two terms, where the convergence rate for the first directly follows the result from \cite{rainforth2018tighter}:
\begin{align}
    \nabla &\mathbb{E} \left[\log \hat{Z}\right]  = \nabla \sum_{t=1}^T \mathbb{E} \left[\log \hat{Z}_t\right] \\
    &= \nabla \log Z -\frac{1}{2K} \Big[\sum_{t=1}^T \nabla \left(\frac{\Var [w_t^1]}{Z_t ^2 } \right) \Big] + \mathcal{O}\left(\frac{T}{K^2}\right) 
\end{align}

For the remaining term that includes the resampling gradient, we apply a thorough analysis as follows.
\begin{align}
    &\mathbb{E} \Big[\nabla \log \prod_{t=2}^T \prod_{k=1}^K \textsc{Categorical}(a_{t-1}^k | w_{t-1}^{1:K}) \cdot \log \hat{Z}\Big]  \nonumber\\
    &= \quad \sum_{t=2}^T \sum_{k=1}^K \mathbb{E} \left[\nabla \log  \textsc{Categorical} (a_{t-1}^k | w_{t-1}^{1:K}) \log \hat{Z}\right] \\
    &=  K \sum_{t=2}^T \sum_{t'=1}^T \mathbb{E} \left[\nabla \log  \textsc{Categorical} (a_{t-1}^1 | w_{t-1}^{1:K})  \log \hat{Z}_{t'}\right] \\
\intertext{Taylor expand $\log\ \hat{Z}_{t'}$ about $Z_{t'}$:}
    &= K \sum_{t=2}^T \sum_{t'=2}^T \mathbb{E} \Bigg\{\nabla \log  \textsc{Categorical} (a_{t-1}^1 | w_{t-1}^{1:K}) \nonumber\\
    &\cdot 
     \Big(\log Z_{t'} + \frac{\hat{Z}_{t'} - Z_{t'}}{Z_{t'}} -\frac{(\hat{Z}_{t'}-Z_{t'})^2}{2Z_{t'}^2} + R_3(\hat{Z}_{t'})\Big) \Bigg\}
\end{align}
where $R_3(\hat{Z}_{t'})$ denotes the remainder in the Taylor expansion of $\log \hat{Z}_{t'}$ about $Z_{t'}$.

For $t'\leq t-1$, we have:
\begin{equation}\label{smallerthan}
\begin{aligned}
     &\mathbb{E} \left[\nabla \log  \textsc{Categorical} (a_{t-1}^1 | w_{t-1}^{1:K}) \cdot \frac{ (\hat{Z}_{t'} -Z_{t'})}{Z_{t'}}\right] \\ 
     & = \mathbb{E}_{\epsilon_{1:t-1}^{1:K}, a_{1:t-2}^{1:K}} 
     \Bigg\{ \frac{\hat{Z}_{t'} - Z_{t'}}{Z_{t'}}  
      \\
     &\quad\quad\times \mathbb{E}_{a_{t-1}^1}\left[ \nabla \log \textsc{Categorical}(a_{t-1}^1 | w_{t-1}^{1:K}) \right] \Bigg\} \\
     &= \mathbb{E}_{\epsilon_{1:t-1}^{1:K}, a_{1:t-2}^{1:K}} \left[ \frac{\hat{Z}_{t'} - Z_{t'}}{Z_{t'}} \cdot 0 \right]=0.
\end{aligned}
\end{equation}

For $t'\geq t$, we have:
\begin{equation}
\begin{aligned}
     & \mathbb{E} \left[\nabla \log  \textsc{Categorical} (a_{t-1}^1 | w_{t-1}^{1:K}) \cdot \frac{ (\hat{Z}_{t'} -Z_{t'})}{Z_{t'}}\right] \\
     &  = \mathbb{E}_{\epsilon_{1:t-1}^{1:K}, a_{1:t-1}^{1:K}} \Bigg\{ \nabla \log \textsc{Categorical}(a_{t-1}^1 | w_{t-1}^{1:K}) 
      \\  &\qquad \qquad\qquad\qquad\qquad \times \mathbb{E}_{\epsilon_{t:t'}^{1:K}, a_{t:t'-1}^{1:K}} \left[
      \frac{\hat{Z}_{t'} - Z_{t'}}{Z_{t'}}\right]
     \Bigg\} \\
    & = \mathbb{E}_{\epsilon_{1:t-1}^{1:K}, a_{1:t-1}^{1:K}} \left[ \nabla \log \textsc{Categorical}(a_{t-1}^1 | w_{t-1}^{1:K}) 
      \cdot 0 \right] \\
    &= 0
\end{aligned}
\end{equation}

Hence, it suffices to compute the convergence rate of the following:
\begin{align}
    K &\sum_{t=2}^T \sum_{t'=2}^T \mathbb{E} \Bigg\{ \\
    &\nabla \log  \textsc{Categorical} (a_{t-1}^1 | w_{t-1}^{1:K}) \cdot \frac{(\hat{Z}_{t'}-Z_{t'})^2}{2Z_{t'}^2}\Bigg\}\nonumber
\end{align}

Note that when $t'\leq t-1$, we obtain similar results as Eq. (\ref{smallerthan}). Thus, we turn to the case when $t' \geq t$.  For $t' \geq t + 1$, each $w_{t'}^k$ has dependence on $a_{t-1}^1$, hence:
\begin{align}
     &K \cdot \mathbb{E} \left[\nabla \log  \textsc{Categorial} (a_{t-1}^1 | w_{t-1}^{1:K}) \cdot \frac{(\hat{Z}_{t'}-Z_{t'})^2}{2Z_{t'}^2}\right] \nonumber\\
     &= K \cdot \mathbb{E} \Bigg\{\nabla \log  \textsc{Categorical} (a_{t-1}^1 | w_{t-1}^{1:K})\nonumber \\
     &\qquad\qquad\qquad\qquad \times \frac{\left(\nicefrac{1}{K} \sum_{k=1}^K(w_{t'}^k-Z_{t'}) \right) ^2}{2Z_{t'}^2}\Bigg\}  \\
     &=  \mathbb{E} \Bigg[\nabla \log  \textsc{Categorical} (a_{t-1}^1 | w_{t-1}^{1:K}) \cdot \frac{(w_{t'}^1-Z_{t'})^2}{2Z_{t'}^2}\Bigg]   \nonumber \\
     \intertext{Applying the score function derivative  trick to the distribution of $a_{t-1}^1$:}
      &= \sum_{i=1}^K \mathbb{E}_{\epsilon_{1:t-1}^{1:K} a_{1:t-2}^{1:K} } \Bigg\{\\
      &\quad\quad\mathbb{E}_{\epsilon_t^1}\left[ \nabla \frac{w_{t-1}^1}{K \hat{Z}_{t-1} } \cdot \frac{(w_t^1 -Z_t)^2}{2Z_t^2} \Bigg|\left[a_{t-1}^1=i\right] \right]  \Bigg\}\nonumber \label{Kterms} \\
     &= K\cdot \mathbb{E}_{\epsilon_{1:t-1}^{1:K} a_{1:t-2}^{1:K} } \Bigg\{\\
     &\quad\quad \mathbb{E}_{\epsilon_t^1}\left[ \nabla \frac{w_{t-1}^1}{K \hat{Z}_{t-1} } \cdot \frac{(w_t^1 -Z_t)^2}{2Z_t^2} \Bigg|\left[a_{t-1}^1=1\right] \right]  \Bigg\}\nonumber\\
     \intertext{Applying the Taylor expansion of $\frac{1}{\hat{Z}_{t-1}}$ around $Z_{t-1}$: $\frac{1}{\hat{Z}_{t-1}} = \frac{1}{Z_{t-1}} + R_2(\hat{Z}_{t-1})$:}
&= \mathbb{E}_{\epsilon_{1:t-1}^{1:K} a_{1:t-2}^{1:K} }\ \Bigg\{\nonumber\\
&\quad\mathbb{E}_{\epsilon_t^1}\left[ \nabla \frac{w_{t-1}^1}{ Z_{t-1} } \cdot \frac{(w_{t'}^1 -Z_{t'})^2}{2Z_{t'}^2} \Bigg|\left[a_{t-1}^1=1\right] \right]  \Bigg\} \nonumber\\
& + \mathbb{E}_{\epsilon_{1:t-1}^{1:K} a_{1:t-2}^{1:K} } \Bigg\{\nonumber\\
&\quad\mathbb{E}_{\epsilon_t^1}\left[ \nabla (w_{t-1}^1 R_2(\hat{Z}_{t-1}))  \cdot \frac{(w_{t'}^1 -Z_{t'})^2}{2Z_{t'}^2} \Big|\left[a_{t-1}^1=1\right]\right]  \Bigg\} 
\end{align}    
    
For $t' = t$, only $w_t^1$ depends on $a_{t-1}^1$,  only one term that conditions on $a_{t-1}^1=1$ in (\ref{Kterms}) is not zero. Consequently we have:
\begin{align}
& K \cdot \mathbb{E} \left[\nabla \log  \textsc{Categorical} (a_{t-1}^1 | w_{t-1}^{1:K}) \cdot \frac{(\hat{Z}_{t'}-Z_{t'})^2}{2Z_{t'}^2}\right] \nonumber\\
&=  \frac{1}{K}\cdot \mathbb{E}_{\epsilon_{1:t-1}^{1:K} a_{1:t-2}^{1:K} } \Bigg\{\nonumber\\
&\quad\mathbb{E}_{\epsilon_t^1}\left[ \nabla \frac{w_{t-1}^1}{ Z_{t-1} } \cdot \frac{(w_{t'}^1 -Z_{t'})^2}{2Z_{t'}^2} \Big|\left[a_{t-1}^1=1\right]\right]  \Bigg\} \nonumber \\
&\quad + \frac{1}{K}\cdot \mathbb{E}_{\epsilon_{1:t-1}^{1:K} a_{1:t-2}^{1:K} } \Bigg\{\nonumber\\
&\quad\mathbb{E}_{\epsilon_t^1}\left[ \nabla (w_{t-1}^1 R_2(\hat{Z}_{t-1}))  \cdot \frac{(w_{t'}^1 -Z_{t'})^2}{2Z_{t'}^2} \Big|\left[a_{t-1}^1=1\right]\right]  \Bigg\} 
\end{align}
    
    \item[2.] Variance.
    \begin{align}
        &\Var \left[\nabla \log \hat{Z}\right] = \Var \Big[\sum_{t=1}^T \nabla\log \hat{Z}_t\Big]\nonumber \\
        &= \sum_{t=1}^T \Var \left[\nabla \log \hat{Z}_t\right] \nonumber\\
        &\quad+ 2  \sum_{t=1}^T \sum_{t' \neq t, t'=1}^T \Cov\left(\nabla \log \hat{Z}_t, \nabla \log \hat{Z}_{t'}\right)
        \label{variance}
    \end{align}
      Decomposing the variance into the sum of variance at each time points, and the pairwise covariance across different time point, we will show that both terms are $\mathcal{O}(\nicefrac{1}{K})$.
      \begin{enumerate}
        \item Variance at each time step.  
        $\forall t=1:T$,
    \begin{align}
        \Var \left[\nabla \log \hat{Z}_t\right] 
        &= \frac{1}{K}\cdot \mathbb{E} \left[\left(\frac{Z_t \nabla w_t^1 - w_t^1 \nabla Z_t}{Z_t^2}\right)^2 \right] \nonumber\\
        &\quad+ \mathcal{O}\left(\frac{1}{K^2}\right) \\
        &= \frac{1}{K}\cdot \mathbb{E} \left[\left( \frac{\nabla w_t^1}{Z_t} \right)^2\right] \nonumber\\
        &\quad+ \mathcal{O}\left(\frac{1}{K^2}\right) \label{varpaper}
        \end{align}

        \item  Covariance between different time steps. 
        
        For $t \neq t' \in 1:T$, we first apply Taylor theorem to $\log \hat{Z}_t$ around $Z_t$, and then exploit the fact that $\hat{Z}_t$ is an unbiased estimation of $Z_t$, and exploit the definition of covariance to expand and collapse terms, as follows:
    \begin{align}
         &\Cov\left(\nabla \log \hat{Z}_t, \nabla \log \hat{Z}_{t'}\right) \\
         &= \Cov \Bigg(\nabla \left(\log Z_t + \frac{\hat{Z}_t - Z_t} {Z_t} + R_1(\hat{Z}_t)\right),\nonumber \\
         &\qquad\qquad \nabla \left(\log Z_{t'} + \frac{\hat{Z}_{t'} - Z_{t'}} {Z_{t'}} + R_1(\hat{Z}_{t'})\right)\Bigg) \nonumber\\
         &= \Cov \Bigg(\nabla \left(\frac{\hat{Z}_t - Z_t} {Z_t} + R_1(\hat{Z}_t)\right),\nonumber\\
         &\qquad\qquad\nabla \left(\frac{\hat{Z}_{t'} - Z_{t'}} {Z_{t'}} + R_1(\hat{Z}_{t'})\right)\Bigg) \\
         &= \mathbb{E} \left[\nabla \left(\frac{\hat{Z}_t } {Z_t}\right) \cdot \nabla \left(\frac{\hat{Z}_{t'} } {Z_{t'}}\right)\right] 
         + \mathbb{E} \left[\nabla \left(\frac{\hat{Z}_{t'}} {Z_{t'}}\right)  \nabla R_1 (Z_t)\right] \nonumber \nonumber\\
         &+ \mathbb{E} \left[\nabla \left(\frac{\hat{Z}_t } {Z_t}\right) \cdot \nabla R_1 (Z_{t'})\right] \nonumber \\
         &+ \Cov \left(\nabla R_1 (\hat{Z}_t), \nabla R_1 (\hat{Z}_{t'})\right)
         \label{var1}
    \end{align}
    
    \begin{enumerate}
  
   \item For the first term in Eq. (\ref{var1}), since $\mathbf{z}_t^{k}$ are i.i.d. for fixed $t$, we have:
    \begin{align}
    &\mathbb{E} \left[\nabla \left(\frac{\hat{Z}_t } {Z_t}\right) \cdot \nabla \left(\frac{\hat{Z}_{t'} } {Z_{t'}}\right)\right] \nonumber \\
    &= \mathbb{E} \left[ \frac{1}{K} \sum_{k=1}^K\nabla \left(\frac{w_t^k}{Z_t} \right) \cdot \frac{1}{K} \sum_{k'=1}^K \nabla \left(\frac{w_{t'}^{k'}}{Z_{t'}}  \right)\right] \\
    &=\ \frac{1}{K^2}\cdot \sum_{k=1}^K \sum_{k'=1}^K \mathbb{E} \left[ \nabla \left( \frac{w_t^k}{Z_t} \right) \cdot \nabla \left( \frac{w_{t'}^{k'}}{Z_{t'}}\right) \right]\nonumber\\
    &= \mathbb{E} \left[\nabla \frac{w_t^1}{Z_t} \cdot \nabla \frac{w_{t'}^1}{Z_{t'}} \right] \\
    &= \Cov \left( \nabla \frac{w_t^1}{Z_t}, \nabla \frac{w_{t'}^1}{Z_{t'}}\right)\label{var2}
    \end{align}

Without loss of generality, we assume $t'> t$. First, when $t'=t+1$, 
  
  \begin{equation}\label{corr}
   \Pr \left(\mathbf{z}_{t+1}^1 \text{ depends on } \mathbf{z}_t^1\right) = \mathbb{E} \left[\frac{w_t^1}{\sum_{k=1}^K w_t^k}\right]  = \frac{1}{K}
  \end{equation}
  
  When $t'>t+1$, using chain rule and by induction we also have,
   \begin{equation}\label{corr}
   \Pr (\mathbf{z}_{t'}^{1} \text{ depends on } \mathbf{z}_t^{1}) = \frac{1}{K}
  \end{equation}
  
  Hence,
  
  \begin{align}
      &\Cov \left( \nabla \frac{w_t^1}{Z_t}, \nabla \frac{w_{t'}^1}{Z_{t'}}\right) \\
      &= \frac{1}{K} \cdot \Cov \left( \nabla \frac{w_t^1}{Z_t}, \nabla \frac{w_{t'}^1}{Z_{t'}} \Bigg|\left[z_{t'}^1 \text{ depends on } z_t^1\right] \right) \nonumber\\
      & \leq \frac{1}{K} \sqrt{ \Var \left[ \nabla \frac{w_t^1}{Z_t}\right] \Var \left[\nabla \frac{w_{t'}^1}{Z_{t'}}\right]} 
\end{align}

  \item For the second and third term in Eq. (\ref{var1}), without loss of generality, we analyze the second term $\mathbb{E} \left[\nabla \left(\nicefrac{\hat{Z}_{t'}} {Z_{t'}}\right) \cdot \nabla R_1 (Z_t)\right]$, and assume $t' > t$.

  Using the i.i.d. property of particles at fixed time step, we have: 
  
  \begin{align}
      &\mathbb{E} \left[\nabla \left(\frac{\hat{Z}_{t'} } {Z_{t'}}\right) \cdot \nabla R_1 (Z_t)\right]\\ &= \frac{1}{K^3}\cdot \mathbb{E} \left[\sum_{k=1}^K \nabla \frac{w_{t'}^k}{Z_{t'}} \mathcal{O} \left(\sum_{k=1}^K (w_t^k-Z_t)^2\right) \right]\nonumber \\
      &= \frac{1}{K}\ \cdot\ \mathbb{E}\left[\nabla \frac{w_{t'}^1}{Z_{t'}}\ \mathcal{O}\left((w_t^1 - Z_t)^2\right)\right]
\end{align}

Similar to the previous analysis on covariance, we can show that

\begin{align}
 \mathbb{E}\left[\nabla\frac{w_{t'}^1}{Z_{t'}} \cdot \mathcal{O}\left((w_t^1 - Z_t)^2\right)\right]&=\mathcal{O}\left(\frac{1}{K}\right)\\
 \intertext{Hence,} 
 \mathbb{E} \left[\nabla \left(\frac{\hat{Z}_{t'} - Z_{t'}} {Z_{t'}}\right) \cdot \nabla R_1 (Z_t)\right] &= \mathcal{O}\left(\frac{1}{K^2}\right)
 \end{align}
 
 \item  For the last term in Eq. (\ref{var1}), note that $ | \Cov (A,B) | \leq  \sqrt{\Var (A)  \Var (B)} $, and $\Var [\nabla R_1 (\hat{Z}_t)] = \mathcal{O}\left(\nicefrac{1}{K^2}\right)$, hence we obtain: 
 \begin{equation}
 \Cov \left(\nabla R_1 (\hat{Z}_t), \nabla R_1 (\hat{Z}_{t'})\right) = \mathcal{O} \left(\frac{1}{K^2}\right)
 \end{equation}
   \end{enumerate}
    Substituting Eq. (\ref{varpaper}), Eq. (\ref{var1}) and Eq. (\ref{var2}) into Eq. (\ref{variance}), we arrive at the final expression for the variance of gradient estimate:
    
    \begin{align}
    &\Var \left[\nabla \log \hat{Z}\right] \nonumber \\
    &= \frac{1}{K} \Bigg\{  \sum_{t=1}^T \mathbb{E} \left[\left( \nabla \frac{w_t^1}{Z_t}\right)^2\ \right] \\
    &\qquad+\sum_{t=1}^T
    \sum_{t'\neq t, t'=1}^T  \sqrt{ \Var \left[ \nabla \frac{w_t^1}{Z_t}\right] \Var \left[\nabla \frac{w_{t'}^1}{Z_{t'}}\right]} \Bigg\} \nonumber \\
    &\qquad+ \mathcal{O}\left(\frac{T^2}{K^2}\right)\nonumber
    \label{finalvar}
    \end{align}    
   \end{enumerate}
   
\end{enumerate}

\end{proof}

\end{document}